\documentclass[letterpaper, 10 pt, conference]{ieeeconf}  % Comment this line out
\IEEEoverridecommandlockouts                              % This command is only
\usepackage[utf8]{inputenc}
\usepackage[T1]{fontenc}
\usepackage{mathptmx} % assumes new font selection scheme installed
\usepackage{mathptmx} % assumes new font selection scheme installed
\usepackage{amsmath} % assumes amsmath package installed

\usepackage{amssymb}  % assumes amsmath package installed
\usepackage{graphicx} 
\usepackage{grffile}
\usepackage[hypcap=true]{caption}
\usepackage{array}
\usepackage{xcolor}
\usepackage{mathtools}
\usepackage{commath}
\usepackage{colortbl}
\usepackage{enumerate}
\usepackage{hyperref}
\usepackage[font=scriptsize]{caption}
\usepackage{algorithmic}
\usepackage{balance}
\usepackage{nopageno} % no page numbers

\title{\LARGE \bf
Stair Climbing using the Angular Momentum Linear Inverted Pendulum Model and Model Predictive Control
}

\author{Oluwami Dosunmu-Ogunbi, Aayushi Shrivastava, Grant Gibson, 
Jessy W Grizzle}

\begin{document}
\maketitle
\thispagestyle{plain}
\pagestyle{plain}

% *********************  BEGIN PAPER  ***********************
% Abstract
\begin{abstract}
\label{sec:abstract}
A new control paradigm using angular momentum and foot placement as state variables in the linear inverted pendulum model has expanded the realm of possibilities for the control of bipedal robots. This new paradigm, known as the ALIP model, has shown effectiveness in cases where a robot's center of mass height can be assumed to be constant or near constant as well as in cases where there are no non-kinematic restrictions on foot placement. Walking up and down stairs violates both of these assumptions, where center of mass height varies significantly within a step and the geometry of the stairs restrict the effectiveness of foot placement. In this paper, we explore a variation of the ALIP model that allows the length of the virtual pendulum formed by the robot's stance foot and center of mass to follow smooth trajectories during a step. We couple this model with a control strategy constructed from a novel combination of virtual constraint-based control and a model predictive control algorithm to stabilize a stair climbing gait that does not soley rely on foot placement. Simulations on a 20-degree of freedom model of the Cassie biped in the SimMechanics simulation environment show that the controller is able to achieve periodic gait.

\end{abstract}

% Introduction/Motivation
\section{Introduction}
\label{sec:intro}

Every day, situations arise that put people's safety and health at risk. As roboticists, we hope that robots will one day offer a means to alleviate some of these risks by taking over dangerous/difficult tasks. Many challenges are preventing us from realizing this hope. One of those is mobility in human-centric spaces.

We live in a world built for bipedal creatures, and thus bipedal robots are a necessary and fundamental addition to a more robot-assisted world. Stairs pose a complicated problem for humans and bipedal robots alike. This paper proposes a method that allows an underactuated bipedal robot to climb a uniform set of stairs. The method employs a variation of the classical inverted pendulum model with a varying center of mass height and a novel combination of virtual constraint-based control and Model Predictive Control (MPC) to achieve a locally exponentially stable stair climbing gait. We first outlined this method in an extended abstract at the \textit{Agile Robotics: Perception, Learning, Planning, and Control Workshop} for the International Conference on Intelligent Robots and Systems in 2022 \cite{ogunbi2022}. This paper expands on the initial presentation.

%%%%%%      BACKGROUND      %%%%%%  
\subsection{Background}
There is a common saying coined by the British statistician George E. P. Box that goes ``all models are wrong, but some are useful.''
In the context of bipedal robotics, roboticists have used a range of models to achieve agile movement in their robots. Full-order dynamical models have proven to be too computationally expensive for practical online control calculations and/or it has proven hard to transfer among different robots of similar morphology. More granular models make it easier to apply a variety of control schemes and perform real-time computations, however, they can also be ineffective in capturing the dominant dynamics of a robot, thus limiting the agility of the closed-loop system (robot plus the controller). In addition, the sim-to-real gap can be hard to manage.

\begin{figure}[b]
    \centering
    \includegraphics[width=0.5\textwidth]{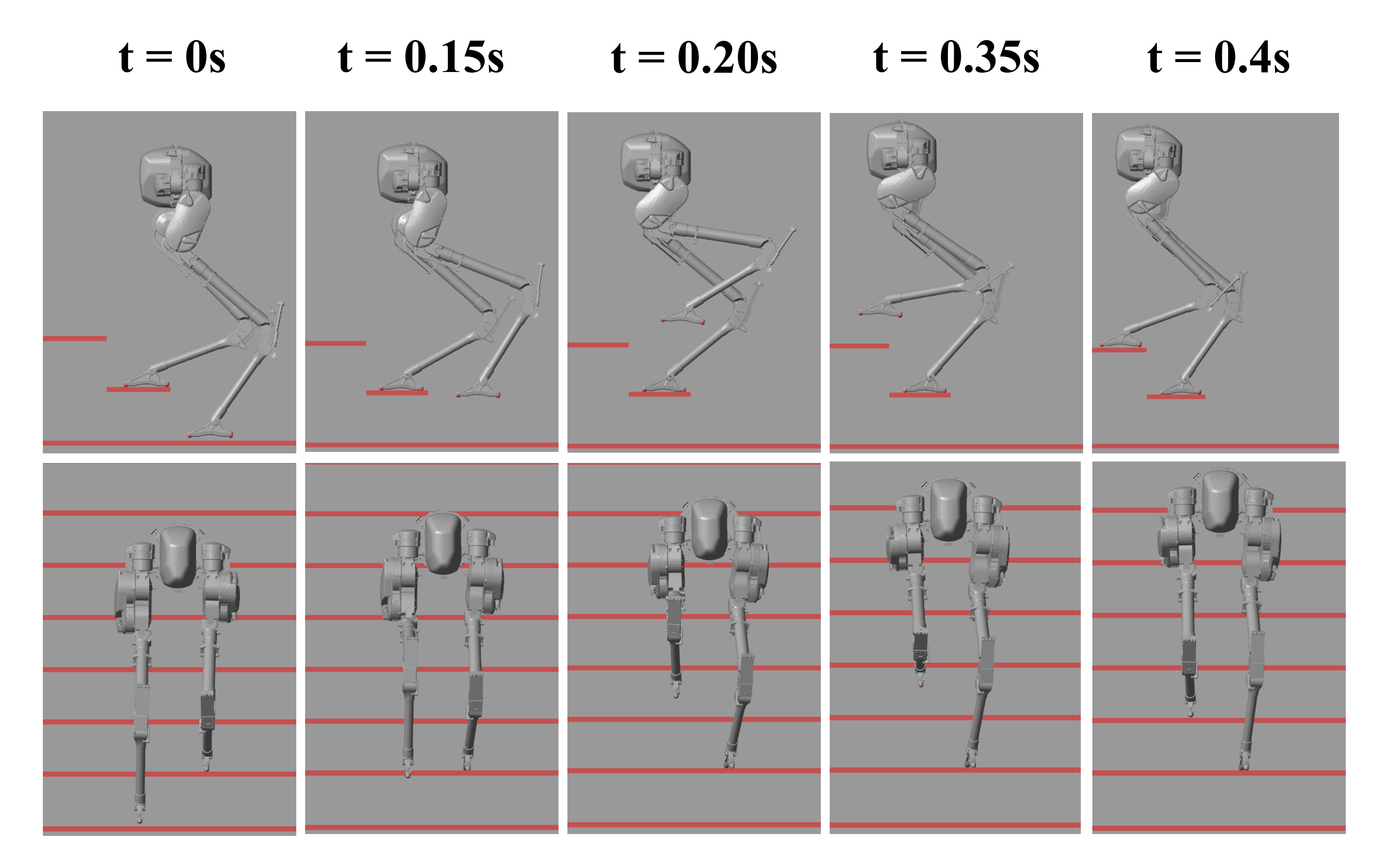}
    \caption{The underactuated Cassie biped walking up stairs in the SimMechanics Simulation environment.}
    \label{fig:cassieSimMechanics}
\end{figure}

The Linear Inverted Pendulum (LIP) model is a popular approach to modeling bipedal locomotion \cite{kajita1991,kajita2001}. The LIP model assumes a point mass fixed on massless legs. Approaches that use the LIP model typically assume a constant center of mass height and use center of mass (CoM) velocity as a means to quantify ``balance'' (e.g., speed stabilization). These assumptions fail to effectively capture impacts associated with gaits where the CoM height undergoes significant variation \cite{powell2016}. Recent research shows that angular momentum about the contact point of the stance foot has higher fidelity when applied to realistic robots \cite{gong2021}. This newer paradigm, called the Angular Momentum Linear Inverted Pendulum (ALIP) model, has been used in control strategies to determine foot placement. Critically, angular momentum  about the support foot has relative degree three with respect to all motor torques except the stance ankle, where it has relative degree one. Consequently, angular momentum  about the support foot is directly controllable via ankle torque and only weakly affected by distal motor torques throughout a step. Furthermore, the transfer of angular momentum property at impact shows that angular momentum about a given contact point is invariant to the impulsive force generated at the contact point.

While the ALIP model has proven to be an effective means of achieving agile locomotion over flat ground \cite{gong2021}, the model has not yet been demonstrated on tasks that involve rapid changes to CoM height such as stair climbing or climbing onto or off objects. Truly agile bipedal robots must be fitted with a controller that is able to handle rapid changes to CoM height to make them capable of navigating cluttered environments.

Model Predictive Control (MPC) is a practical approach to controlling a robot through cluttered environments. By letting the robot ``see ahead of time''--much like humans do when similarly moving through cluttered environments--it is easier to plan control actions that ensure the robot does not fall. The idea of using MPC for bipedal locomotion on non-flat terrain is not new. In \cite{Brasseur2015,jin2021}, the authors generated trajectories for bipedal locomotion on stairs using MPC. Meanwhile, in \cite{Heydari2014}, the authors implemented an MPC-based stair walking controller on a planar robot that had 5 degrees of freedom (DOF).

The study of bipedal robot locomotion over stairs is also not new. Several scholars, such as Fu et al. \cite{Fu2008} and Caron et al. \cite{Caron2019}, have delved into this field by creating stair-walking controllers for fully actuated humanoid robots with 32 and 34 DOF respectively. In \cite{hereid2019}, the authors generated open-loop stair gaits for the 3D underactuated 20 DOF Cassie bipedal robot studied in this report; closed-loop control was not explored. In \cite{verhagen2022}, the authors were able to apply human data of planned and unplanned downsteps on the Cassie biped in simulation. Our paper seeks to further expand the capabilities of the Cassie biped by achieving an asymptotically stable periodic gait on stairs. Prior work by Siekmann et al. \cite{Siekmann2021} made use of reinforcement learning to design a closed-loop controller for the Cassie bipedal robot, perceiving stair height as an unseen perturbation to the controller. Although this achievement is noteworthy, the resulting gait appears to provoke severe impacts, potentially damaging the robot. In this paper, we assume the robot is able to perceive terrain geometry at least one-step ahead, enabling the design of a controller that produces smoother locomotion. Dai et al. \cite{Dai2022} approached the issue by developing a dynamic walking controller for constrained footholds (including on stairs) by regulating an underactuated robot's vertical CoM. We seek an alternative approach to stair climbing using the often-overlooked stance ankle motor.

\subsection{Contributions}
This paper develops a controller that allows the Cassie biped shown in Fig.~\ref{fig:cassieSimMechanics} to climb stairs. Novel contributions include the exploitation of a variation of the ALIP model that allows CoM height to vary within a step, and a novel combination of virtual constraint-based control and MPC to stabilize a stair-climbing gait. 

If the ultimate goal is to have a bipedal robot navigate through cluttered environments, speed may not be the first priority. Rather, precision in balance is a necessity. We show the ability to modulate a robot's closed-loop behavior in real-time so as to smoothly handle stairs as well as reject perturbations on flat ground in SimMechanics simulation.

% Model
\section{Dynamic Model of the Cassie Robot}
\label{sec:model}

The Cassie robot (shown in Fig. \ref{fig:cassieSimMechanics}) is a 32 kg bipedal robot that was designed and built by the company Agility Robotics. Each of its 10 kg legs are actuated at five joints and have two passive joints constrained by springs.

%%%%%%      DYNAMICS      %%%%%%  
\subsection{Floating Base Model}
Bipedal locomotion, such as with stair climbing, can be best characterized using a \textit{hybrid system}--a system that displays both continuous and discrete behavior. The continuous phase describes the dynamics of one foot supporting the robot and the other swinging forward, while the discrete phase describes the transitions between left and right feet. The ``stance leg'' is defined as the leg that is planted on the ground during walking motion. Conversely, the ``swing leg'' refers to the leg whose foot is progressing forward.

Using Lagrangian mechanics, one obtains a second-order differential equation to describe the continuous dynamics for the Cassie biped:
\begin{equation}
D(q)\ddot{q} + C(q,\dot{q})\dot{q} + G(q) = J_{st}^TF + J_s^TF_s + Bu
\label{eq:eqofmotion_unchanged}
\end{equation}
where $D \in \mathbb{R} ^{20 \times 20}$ is the mass inertial matrix, $C \in \mathbb{R}^{20 \times 20}$ is the centrifugal and coriolis forces matrix, $G \in \mathbb{R}^{20 \times 1}$ is the gravitational vector, $J_{st} \in \mathbb{R}^{5 \times 20}$ is the stance foot jacobian (we assume that the blade foot has two points of contact), $F \in \mathbb{R}^{5 \times 1}$ is the ground reaction force acting on the stance foot, $J_s \in \mathbb{R}^{4 \times 20}$ is the jacobian of the springs, $F_S \in \mathbb{R}^{4 \times 1}$ are the forces acting from the springs, $B \in \mathbb{R}^{20 \times 10}$ is the input matrix, $u \in \mathbb{R}^{10 \times 1}$ is the motor torque vector, and $q \in \mathbb{R}^{20 \times 1}$ is the generalized coordinate vector. 

For reasons discussed in the next section, we reformulate the equations of motion defined in \eqref{eq:eqofmotion_unchanged} such that the stance ankle torque term is isolated from the rest of the input terms. Thus,
\begin{equation}
D(q)\ddot{q} + C(q,\dot{q})\dot{q} + G(q) = J_{st}^TF + J_s^TF_s + B_{1} u_{1} + B_{9} u_{9} 
\label{eq:eqofmotion}
\end{equation}
where $B_9 \in \mathbb{R} ^{20 \times 9}$ and $u_9 \in \mathbb{R} ^{9 \times 1}$ are the input matrix and control vector without the stance ankle terms, respectively, and $B_1 \in \mathbb{R} ^{20 \times 1}$ and $u_1 \in \mathbb{R} ^{1 \times 1}$ correspond to the column in the input matrix and value in the control vector relating to the stance ankle torque, respectively.

% Control Philosophy
\section{Control Design Rationale}
\label{sec:Philosophy}
The Cassie biped has 20 DOF to control. This section breaks down how we chose to regulate these degrees of freedom.

During single support (one foot on the ground and the other free of contact), 9 DOF have holonomic constraints imposed on them: four from Cassie's springs (two springs on each leg), and five from the stance foot. Thus, we are left with 11 DOF to control and 10 actuators. The robot is therefore underactuated.

Previous work that has successfully achieved stable walking on level, inclined, and gently rolling terrain consistently used only nine of the ten actuators to achieve stable walking \cite{gong2021zero,gibson2022terrain}, excluding the stance ankle motor. The stance ankle torque is not used in walking because the small ankle motor saturates easily on the real robot in the presence of disturbances, leading to falling. The remaining two uncontrolled degrees of freedom correspond to rotations of the robot about the stance foot in the sagittal and frontal planes and are stabilized via foot placement. As noted by Raibert in \cite{Raibert1984}, if a robot's CoM spends more time in front of the stance foot than behind it, then it generally accelerates, and conversely, it decelerates. This property has been used by many authors to propose foot placement control algorithms \cite{gong2021,redfern1994,Zamani2017,Hong2020,Kanoulas2018,Crews2020,Xiong2021} for stabilization of pendulum models. 

We follow \cite{gong2021zero,gibson2022terrain} and use nine actuators to enforce nine virtual constraints, leaving two degrees of freedom uncontrolled. We adopt the foot placement strategy of \cite{gong2021}  to stabilize the degree of freedom related to rotation about the stance foot in the frontal plane. Stairs offer limited geometry for sagittal foot placement and therefore foot placement in this plane is impractical. Instead, we use intelligent ankle torque control in a manner such that saturation will not destabilize the robot. This is developed in Sec.~\ref{sec:mpc}.

% PBC Controller
\section{Passivity-Based Control}
\label{sec:PBC}
Passivity-Based Control (PBC) is a powerful control strategy used to control nonlinear systems such as bipedal robots \cite{sadeghian2017passivity}. It has practical use for hardware applications because it does not require an accurate model of the system. This is a key feature that adds a layer of robustness to shield from imperfect sensors and uncertain kinematic and dynamic properties within the robot.

We impose a spring constraint such that
\begin{equation}
    J_s \ddot{q} + \dot{J}_s \dot{q} = -K_D^{\text{spring}} J_s \dot{q} - K_P^{\text{spring}} P_s^{\text{error}}
    \label{eq:springConstraint}
\end{equation}
where $P_s^{\text{error}}$ is the spring position error and $K_D^{\text{spring}}$ and $K_P^{\text{spring}}$ are user-defined derivative and proportional controller gains for the springs, respectively.

We additionally impose a non-slip constraint such that
\begin{equation}
    J_{st} \ddot{q} + \dot{J}_{st} \dot{q} = 0.
    \label{eq:nonslip_constraint}
\end{equation}

From \eqref{eq:eqofmotion}, \eqref{eq:springConstraint}, and \eqref{eq:nonslip_constraint} we get
\begin{equation}
    \Tilde{D} f + \Tilde{H} = \Tilde{B}u_9
\end{equation}
where
\begin{equation}
\begin{aligned}
    \Tilde{D} = 
    \begin{bmatrix}
        D       & -J_{st}^\top    & -J_s^\top \\
        J_{st}  & 0            & 0 \\
        J_s     & 0            & 0
    \end{bmatrix}\text{, }
    f =
   \begin{bmatrix}
        \ddot{q} \\
        F_{st} \\
        F_s
    \end{bmatrix}\text{, }
    \Tilde{B} =
    \begin{bmatrix}
        B_9 \\
        0 \\
        0
    \end{bmatrix} \text{, and} \\
   \Tilde{H} =
   \begin{bmatrix}
        C\dot{q} + G - B_1u_1 \\
        \dot{J}_{st} \dot{q} \\
        \dot{J}\dot{q}
    \end{bmatrix}
    - 
    \begin{bmatrix}
        0 \\
        0 \\
        -K_{D}^{\text{spring}} J_s \dot{q} - K_P^{\text{spring}} P_s^{\text{error}}
    \end{bmatrix}.
\end{aligned}
\label{eq:euqtionsofmotionPBC}
\end{equation}

% \begin{equation}
% \begin{aligned}
%     \underbrace{\begin{bmatrix}
%         D       & -J_{st}^\top    & -J_s^\top \\
%         J_{st}  & 0            & 0 \\
%         J_s     & 0            & 0
%     \end{bmatrix}}_{\Tilde{D}}
%     \underbrace{\begin{bmatrix}
%         \ddot{q} \\
%         F_{st} \\
%         F_s
%     \end{bmatrix}}_{f}
%     + \\
%     \underbrace{\begin{bmatrix}
%         C\dot{q} + G - B_1u_1 \\
%         \dot{J}_{st} \dot{q} \\
%         \dot{J}\dot{q}
%     \end{bmatrix}
%     - 
%     \begin{bmatrix}
%         0 \\
%         0 \\
%         -K_{D}^{\text{spring}} J_s \dot{q} - K_P^{\text{spring}} P_s^{\text{error}}
%     \end{bmatrix}}_{\Tilde{H}}
%     =
%     \underbrace{\begin{bmatrix}
%         B_9 \\
%         0 \\
%         0
%     \end{bmatrix}}_{\Tilde{B}}
%     u_9
% \end{aligned}
% \label{eq:euqtionsofmotionPBC}
% \end{equation}

We order the generalized coordinate vector $q$ such that $q = \left[q_c ~~ q_u \right]^\top$, where $q_c$ are the controlled joints and $q_u$ are the uncontrolled joints. We define $\lambda = \left[q_u ~~ F_{st} ~~ F_s \right]^\top$ and partition \eqref{eq:euqtionsofmotionPBC} such that

\begin{equation}
\begin{aligned}
    \Tilde{D}_{11} \ddot{q}_c + \Tilde{D}_{12} \lambda + \Tilde{H}_1 = \Tilde{B}_1 u_9 \\
     \Tilde{D}_{21} \ddot{q}_c + \Tilde{D}_{22} \lambda + \Tilde{H}_2 = \Tilde{B}_2 u_9 . 
\end{aligned}
\end{equation}
That is,
\begin{equation}
    \begin{aligned}
        \begin{bmatrix}
            \Tilde{D}_{11} & \Tilde{D}_{12} \\
            \Tilde{D}_{21} & \Tilde{D}_{22}
        \end{bmatrix}
        \begin{bmatrix}
            \ddot{q}_c \\
            \lambda
        \end{bmatrix}
        +
        \begin{bmatrix}
            \Tilde{H}_1 \\
            \Tilde{H}_2
        \end{bmatrix}
        =
        \begin{bmatrix}
            \Tilde{B}_1 \\
            \Tilde{B}_2
        \end{bmatrix}
        u_9.
    \end{aligned}
\end{equation}

We eliminate $\lambda$ by using Schur Complement, resulting in
\begin{equation}
    \bar{D}\ddot{q}_c + \bar{H} = \bar{B} u_9
\end{equation}
where
\begin{equation}
    \begin{aligned}
        \bar{D} = \Tilde{D}_{11} - \Tilde{D}_{12} \Tilde{D}_{22}^{-1}D_{21} \\
        \bar{H} = \Tilde{H}_1 - \Tilde{D}_{12} \Tilde{D}_{22}^{-1} \Tilde{H}_2 \\
        \bar{B} = \Tilde{B}_1 - \Tilde{D}_{12} \Tilde{D}_{22}^{-1} \Tilde{B}_2.
    \end{aligned}
\end{equation}

We define the output function as
\begin{equation}
    y(x) = h_0(q) - h_d(q, p^{x~des}_{sw}, p^{y~des}_{sw},p^{z~des}_{sw},t)
    \label{eq:y}
\end{equation}
where $h_0$ is the collection of virtual constraints and $h_d$ provides the desired trajectories for the virtual constraints. In part due to precedent \cite{gong2021,gibson2022terrain,gong2019feedback} and in part due to the new ALIP model of Sec.~\ref{sec:alip} that is being used for this paper, the virtual constraints are defined as follows:
\begin{equation}
    h_0(q) = 
    \begin{bmatrix}
    \text{absolute torso pitch}\\
    \text{absolute torso roll}\\
    \text{stance hip yaw}\\
    \text{swing hip yaw}\\
    \text{pendulum length}\\
    p_{st \to sw}^x\\
    p_{st \to sw}^y\\
    p_{st \to sw}^z\\
    \text{absolute swing toe pitch}
    \end{bmatrix}
    \label{eq:h}
\end{equation}
where the pendulum length describes the vector $r_c$ from the stance foot to the CoM and $p_{st \to sw}$ is the vector emanating from the stance foot and ending at the swing foot.

We design a passivity-based controller such that
\begin{equation}
    \bar{D}\ddot{y} + (\bar{C} + K_D) \dot{y} + K_P y = 0
    \label{eq:y_law}
\end{equation}
where $K_D$ and $K_P$ are user-defined derivative and proportional controller gains, respectively. When designing the controller, we check that the decoupling matrix is full rank and we assume that the stance ankle torque is known. The required value of the ankle torque is developed in Sec.~\ref{sec:mpc}.

\section{A Variation of the ALIP Model}
\label{sec:alip}
The ALIP model is a reparameterization of the LIP model where the linear velocity of the CoM is replaced by the angular momentum about the contact point as a key variable to ``summarize'' the state of a robot. For robot models consisting of a single point mass suspended on massless legs, the ALIP model is equivalent to the LIP model. For real robots, with links having distributed mass, reference \cite{gong2021} shows that the ALIP model is superior for making predictions about future state values.

%%%%%%      DERIVATION      %%%%%%  
\subsection{Derivation of the new ALIP Model}

\begin{figure}
    \centering
    \includegraphics[width=0.15\textwidth]{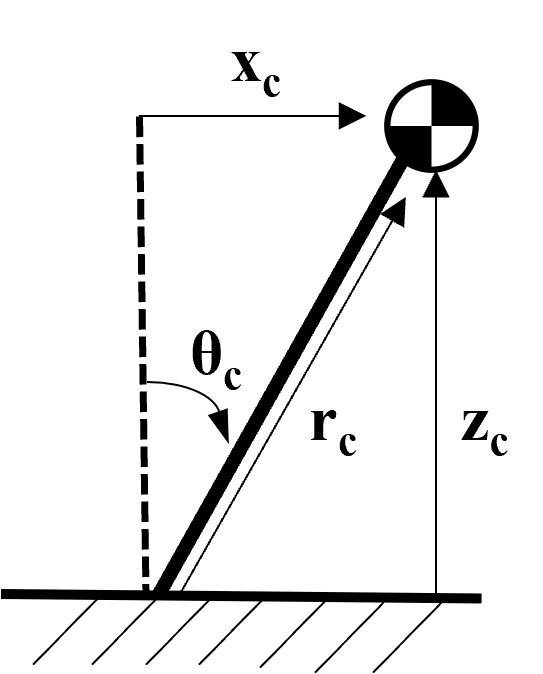}
    \caption{Schematic of an inverted pendulum to derive a variation on the ALIP model.}
    \label{fig:ARIP}
\end{figure}

The derivation of the new ALIP model is as follows. Assume an inverted pendulum as shown in  Fig. \ref{fig:ARIP}, where $(x_c,z_c)$ are the Cartesian position of the CoM with respect to the stance foot. It follows that the angle of the CoM with respect to the stance foot is
\begin{equation}
    \theta_c = \arctan\Big(\frac{x_c}{z_c}\Big).
\end{equation}
Taking the derivative with respect to time yields
\begin{equation}
\label{eq:thetaDot}
    \begin{aligned}
        \dot{\theta}_c &= \frac{1}{1+(\frac{x_c}{z_c})^2} \Big(\frac{\dot{x}_c z_c - \dot{z}_c x_c}{z_c^2}\Big) \\
        &=\frac{1}{z_c^2 + x_c^2}(\dot{x}_c z_c - \dot{z}_c x_c) \\
       & = \frac{1}{r_c^2} (\dot{x}_c z_c - \dot{z}_c x_c)
    \end{aligned}
\end{equation}
where $r_c= \sqrt{x_c^2 + z_c^2}$ is the length of the pendulum. For later use, we rewrite \eqref{eq:thetaDot} as
\begin{equation}
     \dot{\theta}_c = \frac{1}{mr_c^2} (m\dot{x}_c z_c - m\dot{z}_c x_c)
     \label{eq:dthetam}
\end{equation}
where $m$ denotes total mass.

Given the angular momentum about the contact point $L$ and the angular momentum about the CoM, $L_c$, the \textit{angular momentum transfer formula} \cite{gong2021zero} gives
\begin{equation}
    L-L_c = m
    \begin{bmatrix}
    x_c \\
    z_c
    \end{bmatrix}
    \wedge
    \begin{bmatrix}
    \dot{x}_c \\
    \dot{z}_c
    \end{bmatrix}
    = mz_c\dot{x}_c - mx_c\dot{z}_c
    \label{eq:angmomtranafer}
\end{equation}
where
\begin{equation*}
    \begin{bmatrix}
    x_c \\
    z_c
    \end{bmatrix}
    \wedge
    \begin{bmatrix}
    \dot{x}_c \\
    \dot{z}_c
    \end{bmatrix}
    :=
    \begin{pmatrix}
    \begin{bmatrix}
    x_c \\
    0\\
    z_c
    \end{bmatrix}
    \times
    \begin{bmatrix}
    \dot{x}_c \\
    0\\
    \dot{z}_c
    \end{bmatrix}
    \end{pmatrix}
    \bullet
    \begin{bmatrix}
    0\\
    1\\
    0
    \end{bmatrix}.
\end{equation*}
Using \eqref{eq:angmomtranafer}, \eqref{eq:dthetam} becomes
\begin{equation}
    \dot{\theta}_c = \frac{L-L_c}{mr_c^2}.
    \label{eq:dtheta}
\end{equation}

To complete the model, the time derivative of $L$, the angular momentum about the stance leg is
\begin{equation}
    \begin{aligned}
        \dot{L} &= mgx_c  + \tau\\
        &= mg r_c \sin(\theta_c) + \tau,
    \end{aligned}
    \label{eq:dL}
\end{equation}
where $\tau$ is the torque about the contact point, which we will call stance ankle torque. Note that $\tau$ here is equivalent to $u_1$ in \eqref{eq:eqofmotion}. Combining \eqref{eq:dtheta} and \eqref{eq:dL}, the dynamical model becomes
\begin{equation}
\begin{aligned}
    \dot{\theta}_c &= \frac{L-L_c}{mr_c^2}\\
    \dot{L}& = mg r_c \text{sin}(\theta_c) + \tau.
    \end{aligned}
    \label{eq:newALIPv01}
\end{equation}

In \cite{gong2021zero}, it is shown that $L_c$ can be neglected for Cassie-like robots. For the nominal stair climbing trajectory,  $-0.21 \le \theta_c \le 0.13$ radians, and hence we can make the approximation $\sin (\theta_c) \approx \theta_c$. This results in the linear time-varying model
\begin{equation}
\begin{aligned}
    \dot{\theta}_c &= \frac{L}{mr_c^2(t)}\\
    \dot{L}& = mg r_c \theta_c + \tau,
    \end{aligned}
    \label{eq:newALIP}
\end{equation}
which we refer to as the ALIP. The model is time-varying because we will assume that $r_c(t)$ evolves according to the nominal periodic orbit. 

%%%%%%      APPLICATION      %%%%%%  
\subsection{Remarks on the ALIP Model}
When the CoM is controlled to a constant height, the ALIP model becomes linear and time-invariant, and hence admits a closed-form solution. When walking on level ground, a constant CoM assumption renders the impact map linear in the planned horizontal swing foot position.

Walking on stairs violates two of the key assumptions made above: a) the CoM height of the robot must vary to pass from one step to the next, and b) the run of each step of the stair severely restricts horizontal foot placement, effectively eliminating it as a control decision variable. This new version of the ALIP model from \cite{gong2021zero} facilitates accounting for varying pendulum length. We also introduced stance-leg ankle torque into the model so that it can be used as a control variable.

%%%%%%% ACHIEVED Cassie
% \input{Sections/archieve-Cassie}

% MPC
\section{Model Predictive Control using Quadratic Programming}
\label{sec:mpc}
The premise of \textit{Model Predictive Control} (MPC) is to use a \textit{model} of a system to \textit{predict} how the system will evolve over an interval of time to determine an optimal set of \textit{control} inputs to achieve a desired goal state. 

\subsection{Discrete-time Model Formulation} 
We define the state of \eqref{eq:newALIP} to be $x(t) = \left[ \theta_c(t) ~~ L(t) \right]^\top$ and convert the differential equation into a discrete-time model via
\begin{equation}
    \dot{x}(t) \approx \frac{x(t+\Delta t) - x(t)}{\Delta t}.
    \label{eq:dx_approx}
\end{equation}
We let $x_k=x(k \Delta t)$ so that the model can be expressed as 
\begin{equation}
    x_{k+1} = A x_k + b_k u_k
    \label{eq:xkplus1_noB}
\end{equation}
where 
\begin{equation*}
\begin{aligned}
    A_k &=
    \begin{bmatrix}
    1 & 0 \\
    0 & 1
    \end{bmatrix}
    + \Delta t
    \begin{bmatrix}
    0          &    \frac{1}{mr_c(k \Delta t)^2} \\
    mgr_c(k \Delta t)  &    0
    \end{bmatrix} \\
    b_k &=
    \Delta t
    \begin{bmatrix}
    0 \\
    1
    \end{bmatrix}\\
    u_k &= \tau(k \Delta t).
    \end{aligned}
\end{equation*}
While $b_k$ does not vary with time, it is convenient to know which control signal it is distributing in the formulas below. 
Equation \eqref{eq:xkplus1_noB} defines our \textit{model} for MPC. 

\subsection{Predictive Step}
Given our model as well as values for our current state at time $k$, we can calculate the state $k+N$ at the end of a \textit{horizon} of length $N$,
\begin{equation}
\begin{aligned}
    x_{k} & = \text{~given or measured from the robot} \\
    x_{k+1} &= A_{k} x_{k}  + b_{k}  u_{k}  \\
    x_{k+2} &=  A_{k+1} x_{k+1}  + b_{k+1} u_{k+1} \\
    &= A_{k+1} A_{k} x_{k}  + A_{k+1} b_{k}  u_{k} + b_{k+1} u_{k+1} \\
    &~\vdots \\
    x_{k+N} & = A_{k+N-1} \cdots A_{k} x_{k}  +  A_{k+N-1} \cdots  A_{k+1} b_{k}  u_{k} + \\ 
    &~~~~  \cdots A_{k+N-1} b_{k+N-2} u_{k+N-2} + b_{k+N-1} u_{k+N-1}.  \label{eq:stateTransition_derive_init}
\end{aligned}
\end{equation}
For compactness, we rewrite this as
\begin{equation}
    x_{k + N} = S_k x_k + \Gamma_k u_k^{seq}
\end{equation}
where
\begin{equation}
\begin{aligned}
\label{eq:CompactFormula01}
S_k&:=  A_{k+N-1} \cdots A_{k}\\
u^{\rm seq}_k&:=\left[\begin{array}{ccccc} u_{k} & u_{k+1} & \cdots& u_{k+ N-2} &  u_{k+ N-1} \end{array} \right]^\top 
\end{aligned}
\end{equation}
and $\Gamma_k$ can be computed recursively by 
\begin{equation}
\begin{aligned}
B_k & :=b_k\\
B_{k+j} & := \left[ A_{k+j}B_{k+j-1} ~~~ b_{k+j} \right], 1\le j \le N-1\\
 \Gamma_{k} &:=  B_{k+N-1}.
 \end{aligned}
\end{equation}
We note that $\Gamma_{k}$ is a $2 \times N$ matrix.  For $N\ge2$, it can be checked that $\Gamma_k$ is full rank, that is, $\det(\Gamma_k \cdot \Gamma_k^\top) \neq 0$.

With this predictive model, we seek to compute $u^{\rm seq}_k$ such that
\begin{equation}
\label{eq:myMPCequalityConstraint}
    x^{\rm des}_{k+N} = S_k x_k + \Gamma_k u^{\rm seq}_k
\end{equation}
where we'll select $N$ to correspond to the duration of one robot step (i.e., a prediction horizon of 400 ms) and we'll choose $x^{\rm des}_{k+N}$ to be the corresponding value on the nominal periodic orbit at time $t=(k+N)\Delta T,\mod~ T$, where $T=400$ms is the step period.

\subsection{Control Computation}

To minimize the torque sequence $u^{\rm seq}_k$ such that the dynamics hold, we implement a Quadratic Program and arrive at the following optimization problem:

% \begin{equation}
% u^{\rm seq~ \ast}_k:=  \argmin_{ \Gamma_k u^{\rm seq}_k =  \left(x^{\rm des}_{k+N} - S_k x_k \right)} ||u^{\rm seq}_k||^2.
%     \label{eq:minimizationProblem}
% \end{equation}
\begin{equation}\label{eq:QP}
\begin{split}
    &\min_{u_{seq~}}  \Bigg[u^T_{seq~} H(t) u_{seq~} + (x-x^{\text{des}})^\top Q(t) (x-x^{des})\Bigg]\\
    &\mathbf{subject\ to} \\
    & \Gamma_k u^{\rm seq}_k = x^{\rm des}_{k+N} - S_k x_k \\
    & u_{min} < u_k < u_{max}
\end{split}
\end{equation}
where $H(t)$ and $Q(t)$ are weighting matrices, and $u_{min}$ and $u_{max}$ are the lower and upper bounds imposed on the torque input, respectively. We select $H(t)$ and $Q(t)$ such that values toward the end of the step are weighted more, with the value at impacts being weighted the most heavily.

% Lateral Controller (ALIP)
\section{Lateral Stabilization of the Robot}
\label{sec:lateralControl}
We stabilized the lateral motion of the Cassie biped by using the angular momentum-based foot placement strategy developed in \cite{gong2021}, but with the new ALIP model derived in Sec. \ref{sec:alip}.

% Results and Discussion
 \section{Results}
\label{sec:results}
\begin{figure}
    \centering
    \includegraphics[width=0.25\textwidth]{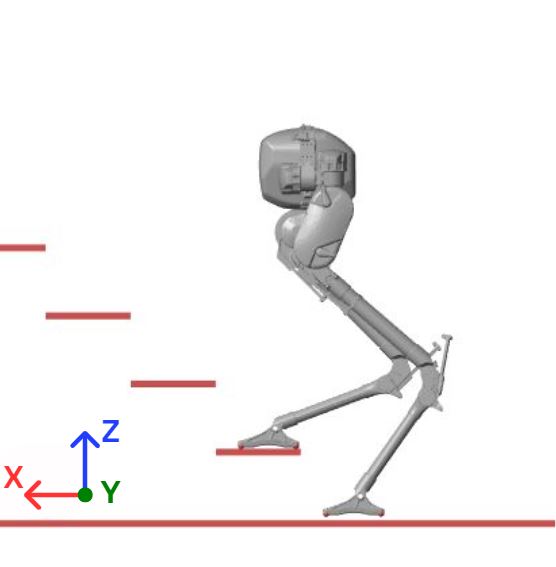}
    \caption{3D model of the Cassie robot in the SimMechanics simulation environment.}
    \label{fig:CassieOnStairsSimulation}
\end{figure}

This section discusses the implementation of the controllers from Sections \ref{sec:PBC}, \ref{sec:mpc} and \ref{sec:lateralControl} on the 20 DOF simulation model of the Cassie robot using Matlab and Simulink. Fig.~\ref{fig:CassieOnStairsSimulation} shows the Cassie robot in the SimMechanics environment on stairs. Note the direction of the positive $x-$ and $z-$axes, which means that a \textit{negative rotation} about the $y-$axis corresponds to \textit{walking up} the stairs. This is an important observation for interpreting later plots.

\subsection{Walking on Flat Ground} 
As a first check, we evaluated our controller on flat ground. We know from previous work \cite{Raibert1984} that foot placement alone on flat ground is enough to stabilize the system. Removing foot placement in the sagittal plane and instead using a fixed step length value (that is, setting the desired swing foot position to a predefined nominal value, similar to what needs to happen on stairs where the sagittal step length is constrained to a constant) results in an unstable closed-loop system. We posited that using ankle torque would then stabilize the system.

Simulations showed this hypothesis to be correct. Turning off ankle torque while the robot walked with fixed step lengths resulted in the robot falling. Adding ankle torque control not only allowed the robot to walk continuously with fixed steps, but also made the system robust against perturbations.
% \jwg{This could be shown by computing the Poincare map for the linear time-varying model. I just note this for future use. You can hide my remark once you read it.} 

\begin{figure}
    \centering
    \includegraphics[width=0.5\textwidth]{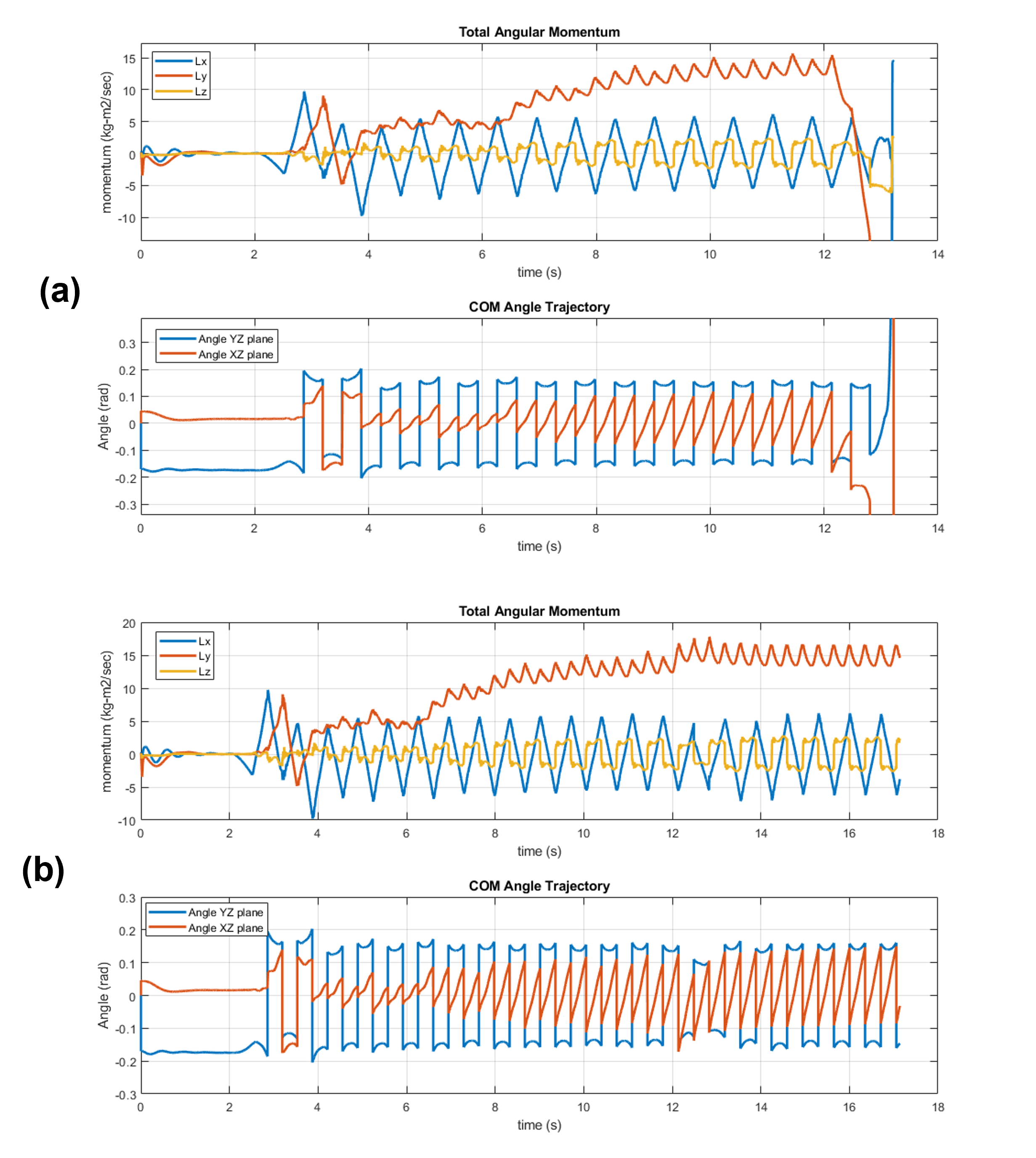}
    \caption{Angular momentum and CoM angle during simulation where robot stands for two seconds, steps in place for the next four seconds, and then is commanded to walk forward at 0.5 m/s for the remainder of the simulation runtime. Fixed step gait is turned on at the 12 second mark. Two test results are shown, (a) not using ankle torque during fixed step, and (b) using ankle torque during fixed step.}
    \label{fig:FixedStepWithAndWithOutAnkleTorqueComparison}
\end{figure}

Fig. \ref{fig:FixedStepWithAndWithOutAnkleTorqueComparison} shows two sets of plots for the total angular momentum and CoM angle for a simulation where the robot stands for the first two seconds, transitions to stepping in place for the next four simulation seconds, and then walks forward for the remainder of the simulation, activating the fixed step gait at the 12-second mark in the simulation time. The first set of plots correspond to the simulation where ankle torque was not used during the fixed step portion of runtime. The second set of plots correspond to the simulation where ankle torque was used during fixed step. Note that the robot falls after just two steps when ankle torque is not engaged during the fixed step gait. This is because the fixed step trajectory does not allow the robot to maintain a periodic angular momentum trajectory, causing the angular momentum to lag behind the desired nominal trajectory and eventually falling. Absent of the intelligent foot placement method that could ensure that a angular momentum trajectory is followed, the system requires a force to maintain stability. Ankle torque supplies this necessary force to the system, pushing the robot back on to the nominal trajectory.

\begin{figure}
    \centering
    \includegraphics[width=0.5\textwidth]{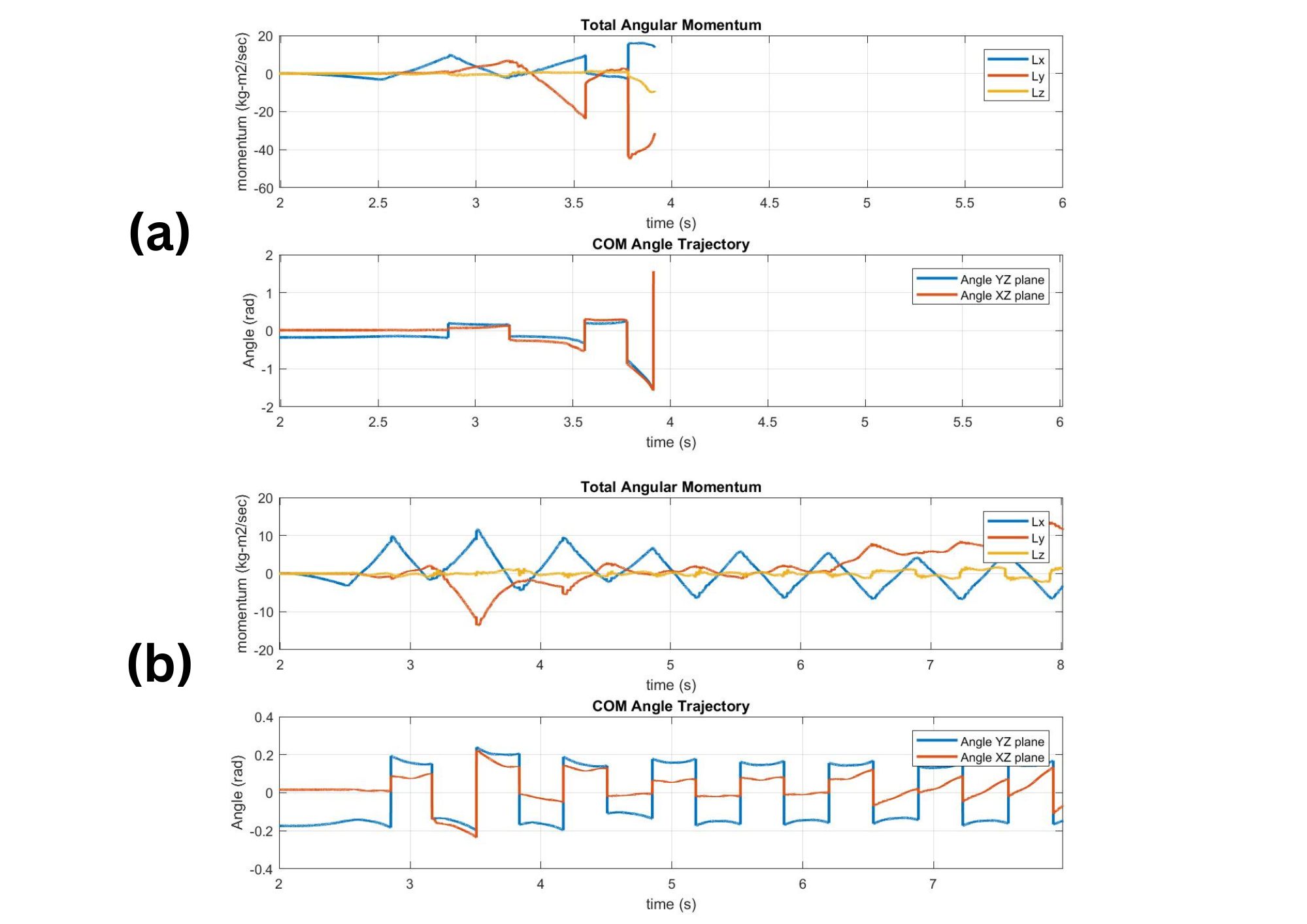}
    \caption{Angular momentum and CoM angle over time  with and without ankle torque to stabilize marching in place with perturbations at $t=3$ sec (a) without ankle torque and (b) with ankle torque. Note, that only the relevant time portion of the plot is shown ($2 < t < 8$) to highlight the effects of the perturbation. In (a), there is no data after $\sim 3.8$ sec because the simulation fails at this time. Data continues until the end of the simulation for (b) because the robot is able to fully recover after the perturbation.}
    \label{fig:perturb_walk_in_place}
\end{figure}

\begin{figure}
    \centering
    \includegraphics[width=0.5\textwidth]{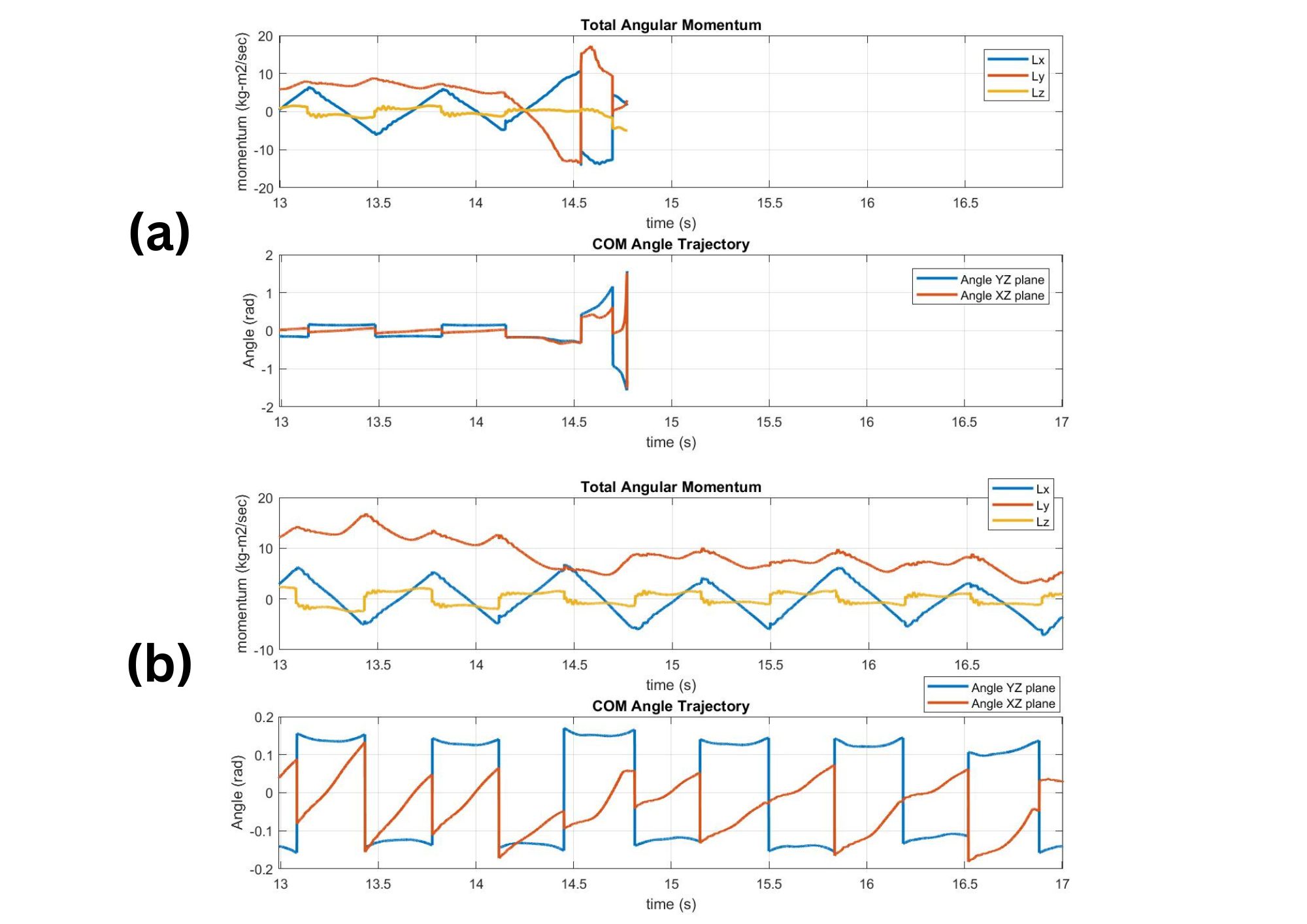}
    \caption{Angular momentum and CoM angle over time with and without ankle torque to stabilize walking forward with perturbations at $t=14$ sec (a) without ankle torque and (b) with ankle torque. Note, that only the relevant time portion of the plot is shown ($13 < t < 17$) to highlight the effects of the perturbation. In (a), there is no data after $\sim 14.7$ sec because the simulation fails at this time. Data continues until the end of the simulation for (b) because the robot is able to fully recover after the perturbation.}
    \label{fig:perturb_walk_forward}
\end{figure}

In Fig.\ref{fig:perturb_walk_in_place} and Fig.\ref{fig:perturb_walk_forward}, we demonstrate the robustness of our ankle torque controller. Following the same gait transitions as aforementioned, we perturb the system at simulation time $t = 3$ seconds (while the robot is walking in place) and $t = 14$ seconds (while the robot is walking forward in fixed steps) by reducing all motor torque inputs by one-fifth ($1/5$) of their desired value for 50 milliseconds. The perturbations resulted in a disturbance equivalent to a shift of 0.1 rad in the CoM angle and 5 kg-m$^2$/sec in angular momentum. In both cases, ankle torque control was able to prevent a fall and return the robot to a periodic gait. In the absence of ankle torque, the robot falls. 

\subsection{Walking up Stairs} 
At each step, the swing foot is regulated to place the new stance foot  near the center of the stair's tread; without this, small errors accumulate and result in the robot not respecting the stair's geometry. In simulations, this is straightforward to achieve. In future experiments, we'll use the perception system design for Cassie in \cite{Huang2023}. 

Using the Passivity Based Controller of Sec.~\ref{sec:PBC} alone to enforce fixed step lengths, without other control in the sagittal plane, resulted in the robot taking two steps and then falling backward. Activating the MPC controller for ankle torque resulted in the 20 DOF simulation model being able to walk an unbounded number of steps. 

\begin{figure}
    \centering
    \includegraphics[width=0.5\textwidth]{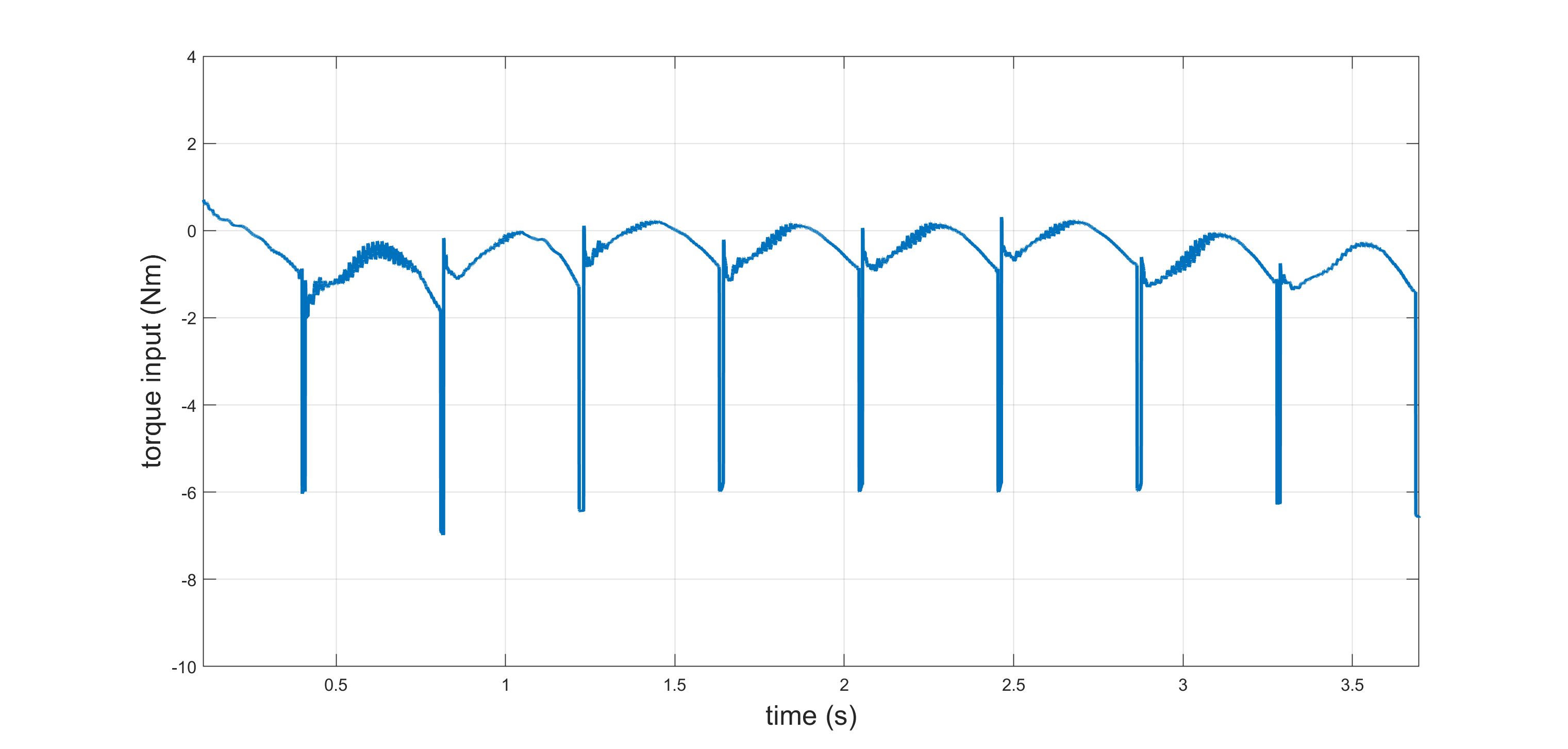}
    \caption{Simulated stance ankle torque vs time using MPC for stair climbing.}
    \label{fig:desiredTorqueInput_mpc}
\end{figure}

Fig. \ref{fig:desiredTorqueInput_mpc} shows the stance ankle torque inputs calculated via the MPC approach throughout the simulation period. We enforced a stance ankle torque limit of $\pm 23$ Nm in the quadratic program solver. This value was decided based on the max torque limit of the ankle motor and the gear ratio of 50. Throughout the simulation, the stance ankle torque is predominantly negative, which means it is ``pushing'' in the direction of motion. Without the additional ankle torque, the robot falls backward, which results in a  positive rotation about the $y-$axis. 

\begin{figure}
    \centering
    \includegraphics[width=0.5\textwidth]{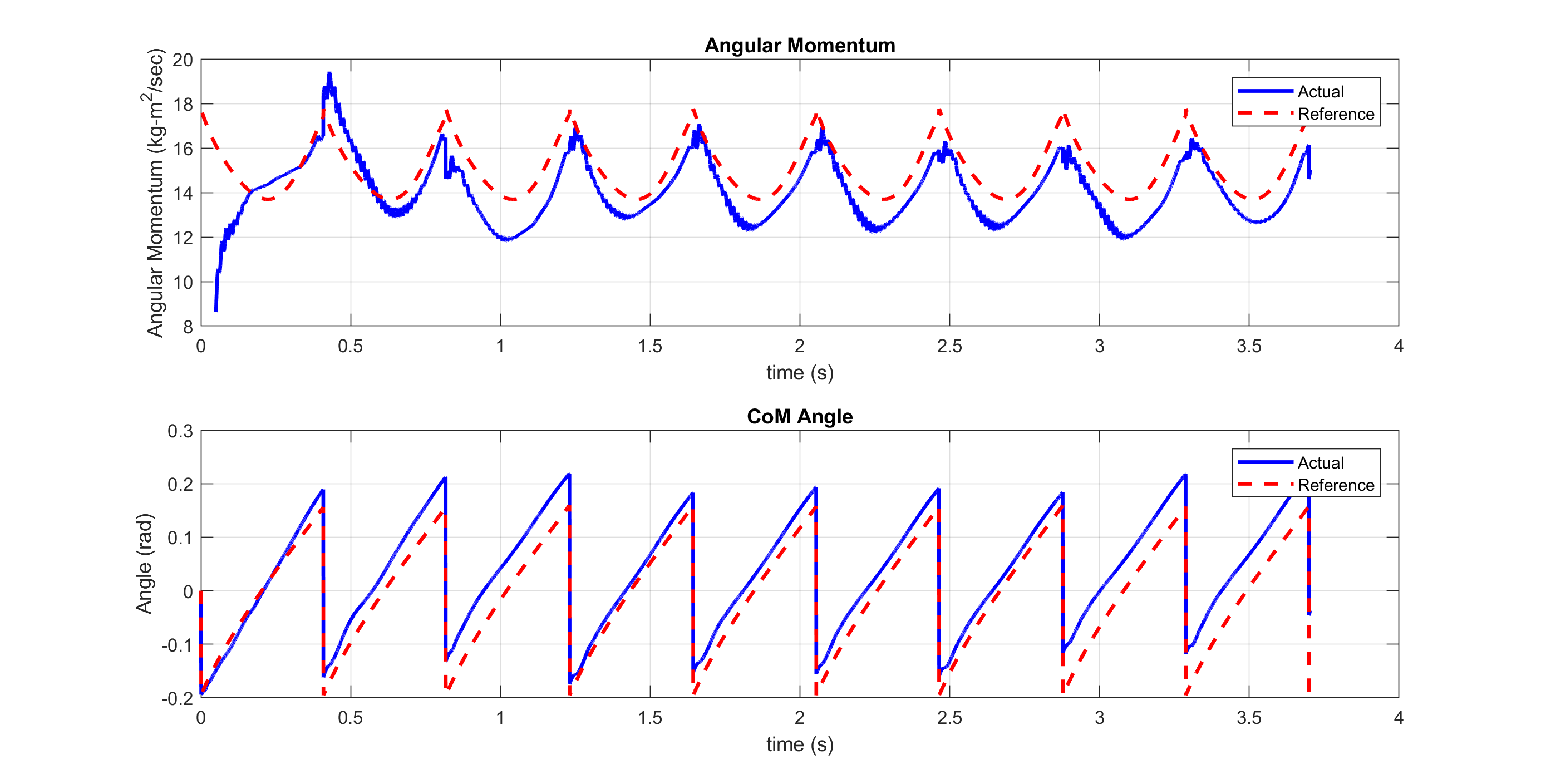}
    \caption{Nominal and simulated angular momentum and CoM angle over time using MPC to determine stance ankle torque to stabilize sagittal motion and (lateral) foot placement to stabilize lateral motion during stair climbing.}
    \label{fig:AngMomAndThetaC_mpc}
\end{figure}

Fig. \ref{fig:AngMomAndThetaC_mpc} shows the angular momentum and CoM angle as the robot walks up 10 stairs. The plots show both the nominal trajectory that was used to set the desired values for the MPC when determining stance ankle torque, as well as the actual simulated values. Note that even though the simulated trajectory is not exactly following the nominal trajectory, it is still able to achieve a stable periodic orbit. The optimized nominal trajectory was developed on a model of the Cassie biped that does not factor in Cassie's springs. We applied our controller on a full order model of the Cassie biped in the SimMechanics simulation environment that includes Cassie's springs as a more faithful representation of the hardware model. Furthermore, we approximate Cassie's states using a Kalman Filter, exactly as we would on hardware, which adds more noise to the system.  In the presence of all of these uncertainties and perturbations, our controller is still able to achieve a stable walking gait up stairs. This is discussed further in the next section.

% %%%%%%%%%%%%%%%%%%%%%%%
\section{Discussion}

The nominal trajectory used for stair climbing was designed with the Fast Robot Optimization and Simulation Toolkit (FROST) \cite{Hereid2017FROST} using a model of Cassie that does not factor in the springs. In effect, the springs, therefore, act as  perturbations to the system that the MPC-generated ankle torque must overcome/accommodate at each impact. 

\begin{figure}
    \centering
    \includegraphics[width=0.5\textwidth]{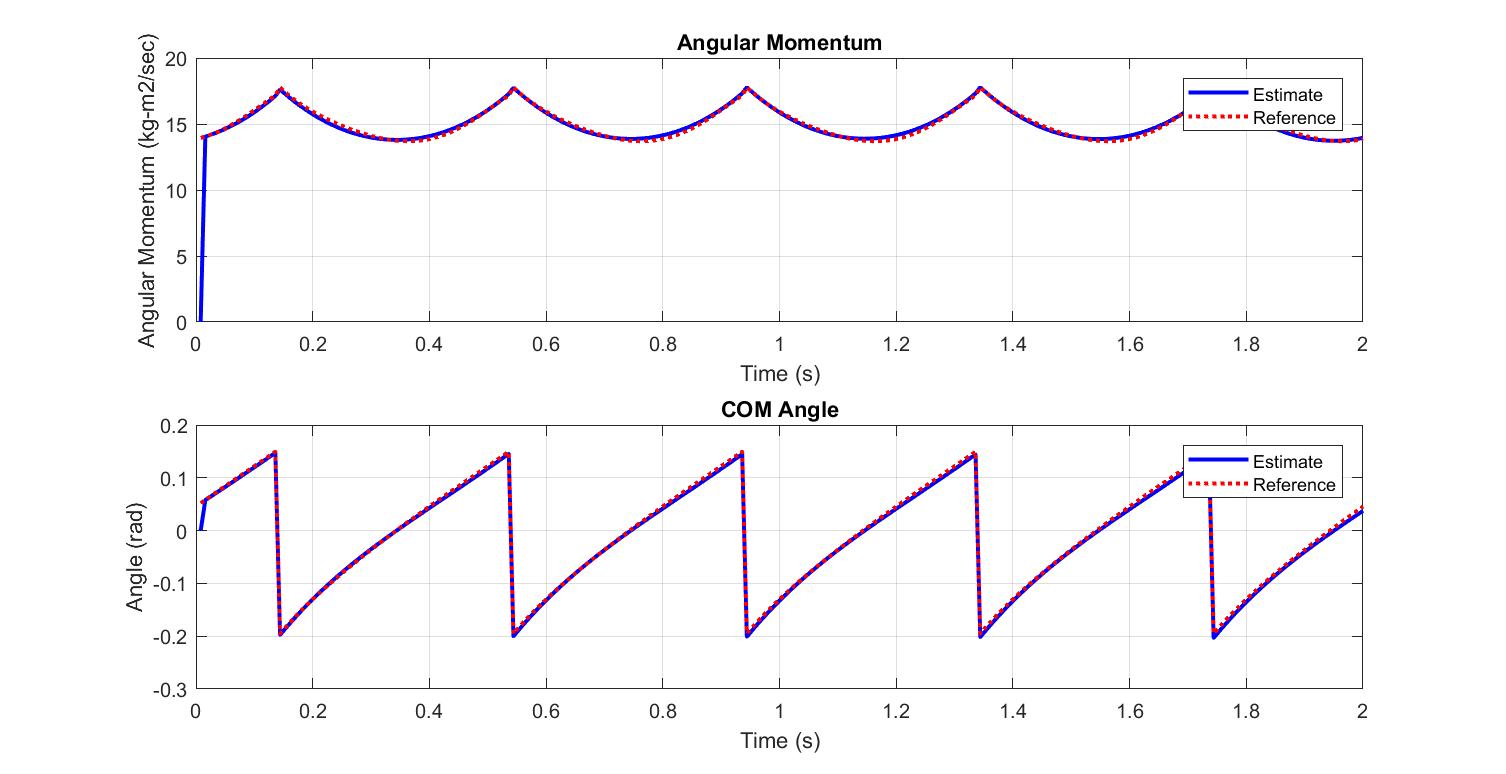}
    \caption{CoM angle and angular momentum over time steps for a horizon length $N=5T$ on the planar nonlinear ALIP model.}
    \label{fig:ThetaCandAngMomentum_NL-ALIP_5stepHorizon}
\end{figure}

\begin{figure}
    \centering
    \includegraphics[width=0.5\textwidth]{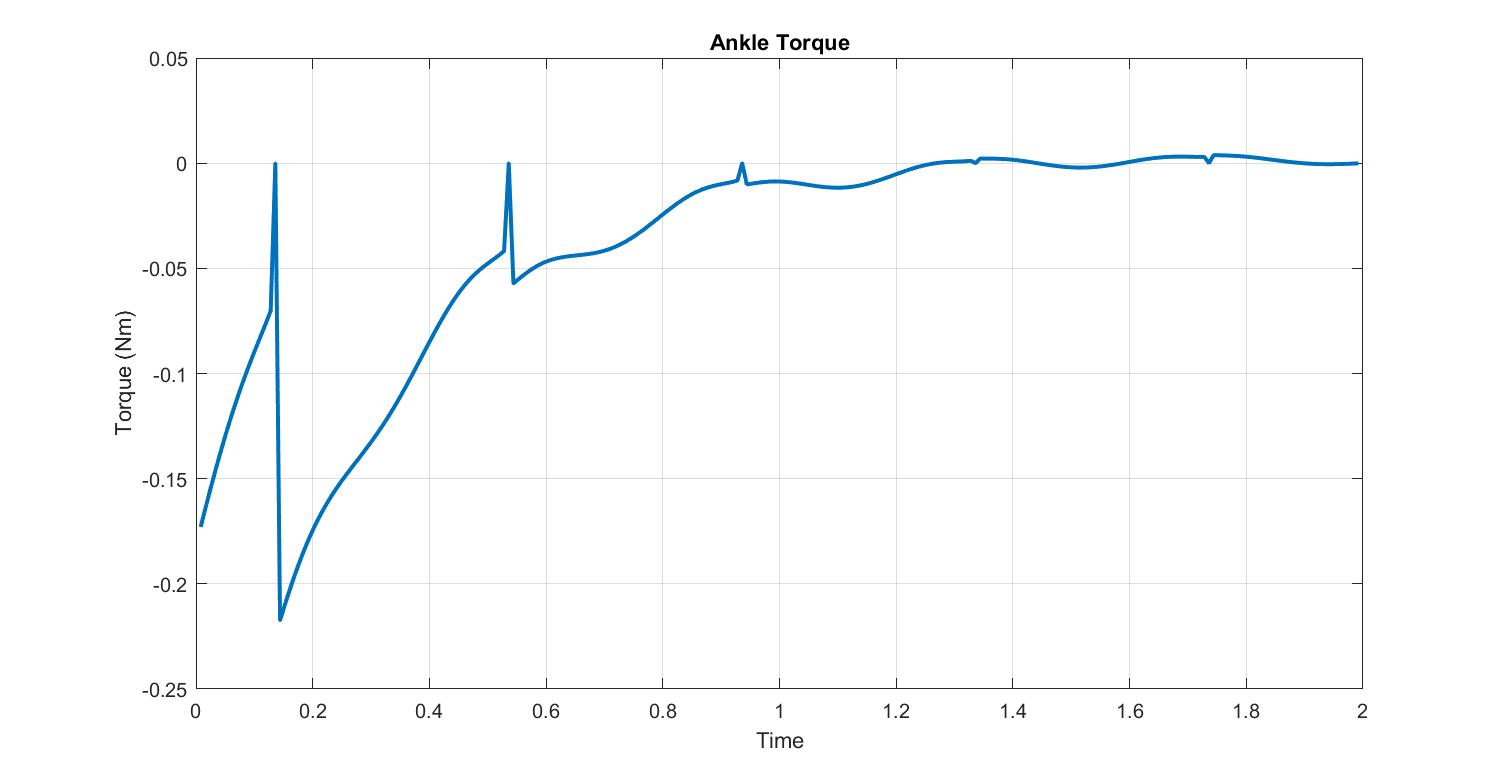}
    \caption{Stance ankle torque over a horizon length $N=5T$ on the planar nonlinear ALIP model.}
    \label{fig:StanceAnkleTorque_NL-ALIP_5stepHorizon}
\end{figure}

At impact, the relatively stiff springs in the stance leg come into play, leading to oscillations in the ``knee joint'' that are not present in the controller design model. This leads to the short-duration spikes in ankle torque seen in Fig.~\ref{fig:desiredTorqueInput_mpc}. To confirm this is the source of the torque spikes, we show in Fig.~\ref{fig:ThetaCandAngMomentum_NL-ALIP_5stepHorizon} a simulation of the planar nonlinear ALIP model in \eqref{eq:newALIP} in closed-loop with the identical controller used on the full-order model of Cassie over a horizon length $N=5T$. As expected, we achieved near perfect tracking with this simplified model compared to the poorer tracking on the full order model shown in Fig. \ref{fig:AngMomAndThetaC_mpc}. Fig \ref{fig:StanceAnkleTorque_NL-ALIP_5stepHorizon} shows the corresponding ankle torques for the simulation on the planar nonlinear ALIP model. Note the marginal torque values that evolve to become almost negligible by the fifth step in the horizon. This matches what we would expect. The optimized trajectories generated by FROST was computed by placing a constraint to minimize stance ankle torque. The planar nonlinear ALIP model is thus able to follow the optimized trajectory using minimal torque input.

While our controller has proven to be robust enough to handle the perturbations caused by the springs, we anticipate that enhanced robustness and agility will require a nominal trajectory that accounts for spring deflection. We can further improve the robustness of our controller by 1) using trajectories that are optimized over a model that factors in Cassie's springs, and 2) upgrading our ALIP model used in MPC to also factor in springs--in effect, using an A-SLIP model. With these changes, our novel control paradigm would not only be able to better handle perturbations to the system during flat ground walking and stair climbing, but also be able to used as the basis of a controller that can help a robot maintain balance while navigating through semi-cluttered environments.

% \jwg{The following discussion seems like a report in a course. The level of sophistication in the discussion needs to be upped.}
% Fig. \ref{fig:AngMomAndThetaC_mpc} shows the angular momentum and CoM angle as the robot walks up 10 stairs. The plots show both the nominal trajectory that was used to set the desired values for the MPC when determining stance ankle torque, as well as the actual simulated values. Note that even though the simulated trajectory is not exactly following the nominal trajectory, it is still able to achieve a stable periodic orbit. This led us to believe that our controllers would be robust enough to reject small perturbations. While the controller was able to reject perturbations for flat ground walking, it was not able to respond in kind for perturbations in stair climbing. Additional investigation is necessary to determine how to improve robustness of the controller to perturbations during stair climbing. However, the fact that ankle torque has proven to increase robustness even while walking on flat ground highlights the promise that this new control paradigm could be used in applications beyond just stairs, helping a robot maintain balance while navigating through semi-cluttered environments.

% Future Work
\section{Conculsions and Future Work}
\label{sec:FutureWork}
We have presented a model-based control strategy for walking up a flight of stairs. The control strategy uses virtual constraints to control the robot's posture. A foot placement strategy ensures lateral stability because standard stair width does not impose any geometric limitations in the lateral direction. In the sagittal plane, however, stair tread depth makes foot placement impractical, and thus we adopted a strategy relying on ankle torque computed via a linearized time-varying model and MPC. Steady-state walking for a 20 DOF simulation model of the Cassie robot was demonstrated in SimMechanics for both flat ground walking and stair climbing. 

The next step will be to apply this strategy on the physical Cassie robot, incorporating a perception system \cite{Huang2023}, so that Cassie is able to navigate stairs autonomously.

\section*{Acknowledgment}
\small{
 Toyota Research Institute
    provided funds to support this work. Funding for J. Grizzle was in part provided by NSF Award No.~2118818.}

% Future Work
% \input{Sections/Future}
%\addtolength{\textheight}{-17cm}

% Appendix
% \begin{appendices}
% knl
% \end{appendices}
% REFERENCE
\nocite{*}
\balance
\bibliographystyle{IEEEtran}
\bibliography{references.bib}

\end{document}